\documentclass[letterpaper]{article} % DO NOT CHANGE THIS
\usepackage{aaai2026}  % DO NOT CHANGE THIS
\usepackage{times}  % DO NOT CHANGE THIS
\usepackage{helvet}  % DO NOT CHANGE THIS
\usepackage{courier}  % DO NOT CHANGE THIS
\usepackage[hyphens]{url}  % DO NOT CHANGE THIS
\usepackage{graphicx} % DO NOT CHANGE THIS
\usepackage{natbib}  % DO NOT CHANGE THIS AND DO NOT ADD ANY OPTIONS TO IT
\usepackage{caption} % DO NOT CHANGE THIS AND DO NOT ADD ANY OPTIONS TO IT
\usepackage{algorithm}
\usepackage{algorithmic}

\usepackage{newfloat}
\usepackage{listings}
\DeclareCaptionStyle{ruled}{labelfont=normalfont,labelsep=colon,strut=off} % DO NOT CHANGE THIS
\floatstyle{ruled}
\newfloat{listing}{tb}{lst}{}
\floatname{listing}{Listing}
\nocopyright
\title{Beyond Hate: Differentiating Uncivil and Intolerant Speech in Multimodal Content Moderation}
\author {
    Nils A. Herrmann\textsuperscript{\rm 1},
    Tobias Eder\textsuperscript{\rm 1},
    Jingyi He\textsuperscript{\rm 1},
    Georg Groh\textsuperscript{\rm 1}
}
\affiliations {
    \textsuperscript{\rm 1}Technical University of Munich\\
    nils\_hermann@outlook.de, tobias.eder@in.tum.de, holly.he@tum.de, grohg@in.tum.de
}

\usepackage{lipsum}
\usepackage{tabularx}
\usepackage{booktabs}
\usepackage{makecell}
\usepackage{multirow}

\usepackage[dvipsnames]{xcolor}
\usepackage{xstring}

\newcommand{\autometric}[2]{%
  \IfBeginWith{#2}{-}
    {#1\quad{\scriptsize\textcolor{BrickRed}{(#2)}}}%
    {#1\quad{\scriptsize\textcolor{ForestGreen}{(#2)}}}%
}

\begin{document}

\maketitle

\begin{abstract}
Current multimodal toxicity benchmarks typically use a single binary hatefulness label. This coarse approach conflates two fundamentally different characteristics of expression: tone and content. Drawing on communication science theory, we introduce a fine-grained annotation scheme that distinguishes two separable dimensions: \textit{incivility} (rude or dismissive tone) and \textit{intolerance} (content that attacks pluralism and targets groups or identities) and apply it to 2,030 memes from the Hateful Memes dataset. We evaluate different vision-language models under coarse-label training, transfer learning across label schemes and a joint learning approach that combines the coarse hatefulness label with our fine-grained annotations. Our results show that fine-grained annotations complement existing coarse labels and, when used jointly, improve overall model performance. Moreover, models trained with the fine-grained scheme exhibit more balanced moderation-relevant error profiles and are less prone to under-detection of harmful content than models trained on hatefulness labels alone (FNR-FPR, the difference between false negative and false positive rates: 0.74 to 0.42 for LLaVA-1.6-Mistral-7B; 0.54 to 0.28 for Qwen2.5-VL-7B). This work contributes to data-centric approaches in content moderation by improving the reliability and accuracy of moderation systems through enhanced data quality. Overall, combining both coarse and fine-grained labels provides a practical route to more reliable multimodal moderation.
\end{abstract}

\section{Introduction}
Content moderation on online platforms increasingly relies on automated systems to detect harmful multimodal content such as memes\footnote{\textbf{Content warning:} This paper contains images and text taken from the Hateful Memes dataset, which contains hateful and abusive language, identity-based slurs, and offensive stereotypes, including content targeting individual groups (e.g., race, ethnicity, religion, gender, sexual orientation). It may also include harassment and references to violence.} \cite{martinezpandianiToxicMemesSurvey2025}. These systems are trained on benchmark datasets whose label definitions implicitly encode normative choices about what counts as harmful. The most widely used multimodal benchmark, the Hateful Memes Dataset \cite{kielaHatefulMemesChallenge2020}, uses a single binary distinction between hateful and benign content. This coarse operationalization necessarily conflates two fundamentally different characteristics of expression, namely tone and content.

This conflation is not just a technical inconvenience. From a normative perspective, communication science research has established that incivility - rude or dismissive tone - and intolerance - content that attacks pluralism and targets identities - are conceptually and empirically distinct \cite{coeOnlineUncivilPatterns2014, rossiniIncivilityUnderstandingPatterns2022, kumpelDifferentialPerceptionsReactions2023}. Uncivil speech may still contribute to democratic deliberation despite its tone \cite{rossiniIncivilityUnderstandingPatterns2022}, while intolerant speech poses a fundamentally different kind of threat by seeking to dehumanize or exclude. A moderation system that cannot distinguish between these dimensions risks both over-moderating legitimate disagreement and under-detecting harmful content that happens to be expressed in civil language. Existing multimodal benchmarks, by collapsing these dimensions into a single label, obscure this distinction and propagate it into every model trained on their data.

In practice, using a fine-grained \textit{multi-label approach} which distinguishes between tone (incivility) and content (intolerance) enables the development of nuanced annotation schemes, which in turn lead to more reliable data with versatile use-cases. Drawing on the theoretical distinction between incivility and intolerance, we introduce a fine-grained annotation scheme and apply it to 2,030 memes from the Hateful Memes dataset. We evaluate vision-language models under coarse-label training, transfer learning across label schemes, and a joint learning approach that combines the coarse hatefulness label with our fine-grained annotations. We situate our work across unimodal, multimodal, and generative approaches that have been proposed for toxic meme detection \cite{waseemUnderstandingAbuseTypology2017, hossainIntermodalAttentionFramework2023, liangTRICANMultiModalHateful2022, kumarHateCLIPperMultimodalHateful2022, quEvolutionHatefulMemes2023, koutlisMemeFierDualstageModality2023, linSurfaceUnveilingHarmful2023, linExplainableHarmfulMeme2024, caoPromptingMultimodalHateful2022, jainGenerativeModelsVs2023}.

Our results show that the original binary hatefulness label captures most forms of identity-based intolerance reliably, but largely misses politically-oriented intolerance and the dimension of incivility. Moreover, we show that automated models trained with fine-grained labels exhibit more balanced moderation-relevant error profiles: Open-source VLMs shift from systematic under-detection of harmful content towards symmetric error trade-offs (FNR-FPR: 0.74 to 0.42 for LLaVA-1.6-Mistral-7B; 0.54 to 0.28 for Qwen2.5-VL-7B), a finding with direct implications for platforms deploying these models for content moderation.

\paragraph{Research Questions.}

To evaluate whether a conceptually grounded distinction between tone and content improves multimodal moderation models, we address the following four research questions:
\begin{enumerate}
  \item \textbf{RQ1 (Construct separability):} Do incivility (tone) and intolerance (content) empirically separate in multimodal memes, and how often do they co-occur?
  \item \textbf{RQ2 (Predictive utility):} Does supervision with fine-grained labels (incivility and intolerance), alone or jointly with the original coarse hatefulness label, improve model performance compared to coarse-only supervision?
  \item \textbf{RQ3 (Error trade-offs):} How does label granularity affect moderation-relevant error trade-offs when detecting harmful content?
  \item \textbf{RQ4 (Operationalization alignment):} What does the original binary hatefulness label operationalize in practice: intolerance, incivility, their conjunction, or their disjunction?
\end{enumerate}

\paragraph{Contributions.}
We make four contributions:
\begin{enumerate}
    \item \textbf{Fine-grained annotation scheme:} We introduce a multimodal annotation scheme that operationalizes harmful expression along two dimensions: \emph{incivility} (tone) and \emph{intolerance} (content). The scheme is grounded in communication science and accompanied by detailed guidelines and sub-types to support consistent labeling.
    \item \textbf{Open-access dataset and code:} We release an annotated subset of the Hateful Memes dataset containing fine-grained incivility and intolerance labels (in compliance with the dataset license), along with annotation guidelines, documentation, and scripts that allow users to reconstruct the full dataset.
    \item \textbf{Empirical evaluation of label granularity:} We evaluate vision-language models under four training configurations (zero-shot, baseline fine-tuning, transfer learning and joint learning) to quantify how coarse versus fine-grained supervision affects learnability and generalization across tasks.
    \item \textbf{Moderation-relevant error and construct analysis:} We analyze false positive and false negative behavior to assess whether different supervision schemes bias models toward over-flagging or under-detection. Additionally, we test alternative operationalizations linking the original binary hateful label to incivility and intolerance, clarifying what single-label hate detection captures in practice.
\end{enumerate}

\section{Related Work}
\subsection{Current Multimodal Computational Approaches}
Effectively identifying hateful memes requires multimodal understanding, as the meaning often emerges from the interaction between visual and textual components. Unimodal models which process only text or image, struggle to capture these interactions, particularly in the presence of benign confounders where neither modality alone is sufficient \cite{kielaHatefulMemesChallenge2020}.

Two major computational paradigms have emerged for this task. The first is the \textit{specialist approach}, which relies on CLIP-style encoders combined with task-specific fusion and interaction mechanisms for multimodal hate classification \cite{radfordLearningTransferableVisual2021}. Notable examples include HateCLIPper \cite{kumarHateCLIPperMultimodalHateful2022} and MemeCLIP \cite{shahMemeCLIPLeveragingCLIP2024}. The second is the \textit{generalist approach}, which employs instruction-tuned vision-language models such as LLaVA \cite{liuVisualInstructionTuning2023} and Qwen2-VL \cite{wangQwen2VLEnhancingVisionLanguage2024}, adapting them to the benchmark via supervised fine-tuning and related strategies. More recent work augments such models with retrieval or staged adaptation procedures to increase robustness \cite{meiRobustAdaptationLarge2025}. Together these two modeling families provide a natural basis for studying how annotation choices interact with multimodal learning.

\subsection{Datasets for Hateful Meme Identification}

Most multimodal hate speech datasets, including the \textit{Hateful Memes Challenge} \cite{kielaHatefulMemesChallenge2020}, employ a single binary label to distinguish hateful from non-hateful content \cite{sabatHateSpeechPixels2019, badourHatefulMemesClassification2021}. 

Subsequent work has introduced more nuanced labeling schemes. \citet{rajputHateMeNot2022} frame hatefulness as a multi-class problem, differentiating between hate-inducing, satirical, and non-offensive memes, while \citet{bhandariCrisisHateMMMultimodalAnalysis2023} extend single-label approaches by distinguishing between directed and undirected hate.

Other studies conceptualize hatefulness as a multi-dimensional phenomenon captured through multiple labels. \citet{groverPoliMemeExploringOffensive2025} distinguish between the sentiment, content, direction, and target of toxic memes. \citet{linGOATBenchSafetyInsights2025} introduce GOAT-Bench, which covers related dimensions such as hatefulness, misogyny, offensiveness, sarcasm, and harmfulness, with each dimension based on a different dataset and treated as a separate classification task. \citet{kumariEmoffMemeIdentifyingOffensive2023} show that jointly learning offensiveness with related dimensions improves detection performance, further indicating that hatefulness cannot be captured by a single label.

Research most closely related to this study augments the original Hateful Memes dataset with additional annotation layers. \citet{mathiasFindingsWOAH52021} expand the dataset by labeling protected categories (e.g., race, religion, disability) and attack types (e.g., mocking, dehumanizing, inciting violence), while \citet{heeDecodingUnderlyingMeaning2023} add contextual explanations for why content is considered hateful. Collectively, these efforts provide a more detailed view of multimodal hatefulness. However, none jointly annotate tone and content, leaving open questions about how stylistic and semantic elements co-occur in multimodal expressions of hate.

\subsection{Multi-label Hatefulness}
There are recent efforts to disentangle different forms of harmful online expression by distinguishing between incivility and intolerance. \citet{rossiniIncivilityUnderstandingPatterns2022} argues that these two forms of speech are not only conceptually distinct but also empirically separable in political discourse. Incivility, marked by rudeness, often emerges in discussions that reflect the vibrant, even confrontational, nature of democratic deliberation. By contrast, intolerance refers to speech that undermines democratic values by attacking rights, identities, or pluralism. \citet{rossiniIncivilityUnderstandingPatterns2022} provides evidence that incivility can occur even in productive political engagement, while intolerance appears in contexts that exacerbate harm. It is crucial to recognize this distinction in order to reframe debates about moderation, which differentiate between an uncivil tone and the more substantial threats posed by intolerant content.

Our approach builds on \citet{bianchiItsNotJust2022}, who apply a multi-label framework to a large corpus of immigration-related tweets from the US and UK. They use text-based models to detect both incivility and intolerance. This multi-label approach improves understanding of harmful discourse, producing models that outperform those trained on coarse-grained datasets. These results reinforce the idea that multi-label schemes can improve both the conceptual clarity and the empirical performance of content moderation.

\section{Data}
\subsection{Hateful Memes Dataset}

\begin{figure}[ht]
    \centering
    \includegraphics[width=\linewidth]{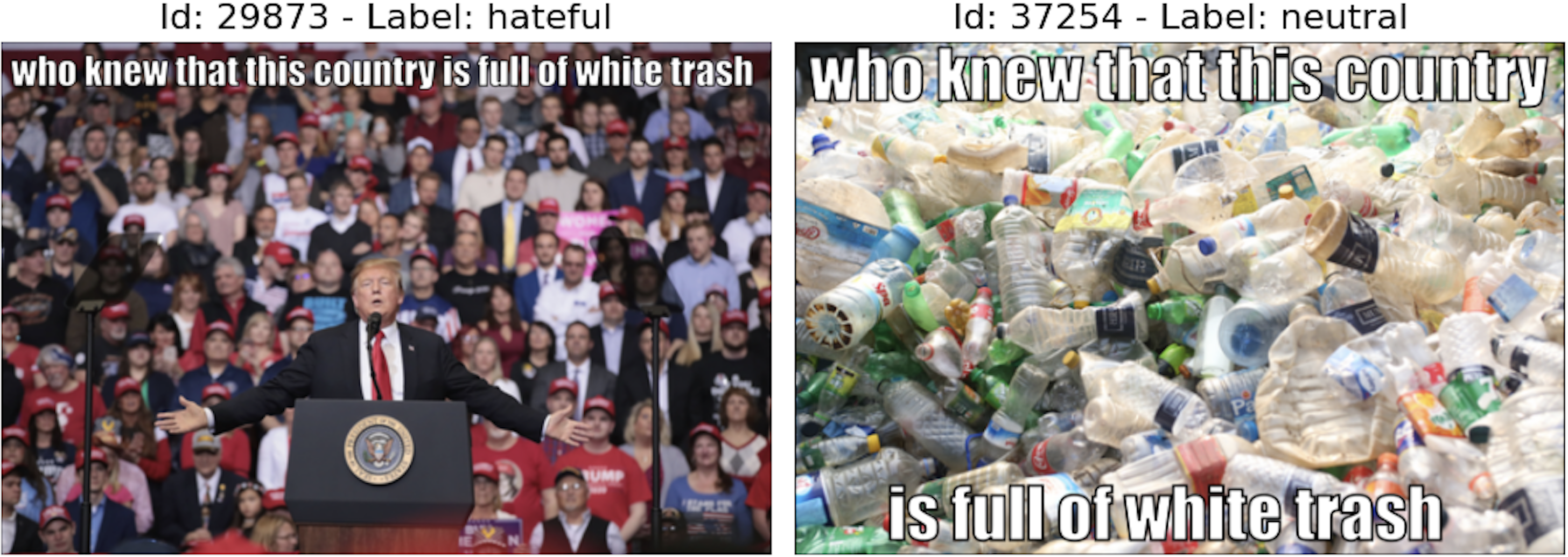}
    \caption{Example memes from the hateful meme dataset that illustrate the concept of 'benign confounders'. Image above is a compilation of assets, including ©Getty Images.}
    \label{fig:memes}
\end{figure}

The Hateful Memes dataset \cite{kielaHatefulMemesChallenge2020} was introduced as a technical benchmark for multimodal hate speech detection. Its goal is to test whether models can effectively combine image and text information to detect hateful content. The dataset includes 'benign confounders' to make the task more challenging and avoid shortcuts based on unimodal cues. Figure \ref{fig:memes} illustrates the concept of benign confounders, which are cases where neither the image nor the text alone is sufficient to determine whether content is hateful.

The dataset contains approximately 10,000 memes, each consisting of an image paired with embedded text. Labels are binary: memes are either marked as hateful or not. The challenge defines hateful content as follows:
\begin{quote}
    A direct or indirect attack on people based on characteristics, including ethnicity, race, nationality, immigration status, religion, caste, sex, gender identity, sexual orientation, and disability or disease. We define attack as violent or dehumanizing (comparing people to non-human things, e.g. animals) speech, statements of inferiority, and calls for exclusion or segregation. Mocking hate crime is also considered hate speech.
\end{quote}
Although previous research \cite{davidsonAutomatedHateSpeech2017} proposed a three-way distinction separating hateful, offensive, and normal content, the Hateful Memes challenge opted for a single binary label. They justify this choice by arguing that the distinction between hatefulness and offensiveness is ``murky, and the classification decision less actionable'' \cite[p.~3]{kielaHatefulMemesChallenge2020}.

While a single binary label is appropriate for a technical benchmark that prioritizes standardized evaluation over realism \cite{orrAISportCompetitive2024}, real-world moderation demands finer distinctions. \citet{kernAnnotationSensitivityTraining2023} show that annotation clarity and agreement are highest when annotators classify tone and content simultaneously, underscoring the practical value of separating these dimensions. Omitting incivility also deprives models of linguistic cues that can reveal cases where tone amplifies or conceals harmful content.

\subsection{Annotation of the Hateful Memes Dataset}
To introduce a more nuanced understanding of hatefulness, we annotate a subset of the Hateful Memes dataset with two variables that distinguish between incivility and intolerance. Specifically, we annotate a 21\% stratified random sample by split of the original dataset, corresponding to 2030 datapoints.\footnote{The fine-grained subset annotation labels will be made available online. The annotations are provided in Parquet format with documented schemas and an explicit license, ensuring FAIR compliance \cite{fair}.} Our goal is representativeness with respect to the Hateful Memes benchmark. The annotated subset is intended to approximate the content and label distributions of each split, rather than to represent online meme distributions in the wild. In addition to representativeness, we chose this sample size to be sufficient for both reliable descriptive and associative analyses of label co-occurrence and for supervised adaptation of the evaluated models. Empirically, learning-curve experiments reported in the Appendix show that, for our fine-tuned VLMs, performance on the fine-grained tasks saturated well before the full 2,030 samples, indicating diminishing returns from annotating additional items. We therefore treat the 21\% subset as an adequate compromise between annotation cost and model reliability.

Our annotation scheme is grounded in definitions from communication science, drawing particularly on the work of \citet{rossiniIncivilityUnderstandingPatterns2022} and \citet{coeOnlineUncivilPatterns2014}. Incivility and intolerance are treated as two separate variables, each capturing a different dimension of discourse.

\textbf{Incivility} is defined as
\begin{quote}
    Rude, disrespectful or dismissive \textit{tone} towards others as well as opinions expressed with antinormative intensity.
\end{quote}
\textbf{Intolerance} is defined as
\begin{quote}
    \textit{Content} that is threatening to democracy and pluralism—such as prejudice, segregation, hateful or violent speech, and the use of stereotyping in order to disqualify others and groups.
\end{quote}

Annotators are provided with definitions of incivility and intolerance, along with lists of their respective sub-types. The scheme specifies three types of incivility and nine types of intolerance. A full description and definition of the different subtypes can be found in the Appendix. Annotators must indicate the specific sub-type whenever any is present. For fine-tuning and evaluation, the annotations are consolidated into two separate binary labels. The binarization preserves the core incivility-intolerance distinction while remaining compatible with the original benchmark. Subtype labels are retained in the release data and used for annotation training, further analysis and adjudication.

Table \ref{tab:annotation-types} contrasts the two annotation strategies used in this study. The original approach applies a coarse single-label scheme, classifying content as either \emph{hateful} or \emph{neutral}. Our approach adopts a fine-grained multi-label scheme with two binary labels, distinguishing \emph{intolerant} from \emph{tolerant} and \emph{uncivil} from \emph{civil}.

\begin{table}[H]
\centering
\small
\begin{tabular}{l l c c}
\toprule
\textbf{\makecell{Annotation \\ type}} & \textbf{Labels} &  Count  & \makecell{\% of original \\ dataset} \\
\midrule
\makecell[l]{Coarse \\ single-label} &
\makecell[l]{Hateful vs. Neutral} & 9664 & 100\% \\
\midrule
\makecell[l]{Fine-grained \\ multi-label} & 
\makecell[l]{Intolerant vs. Tolerant \\ Uncivil vs. Civil} & 2030 & 21\% \\
\bottomrule
\end{tabular}
\caption{Two annotation schemes used in this study, with their respective coverage in the dataset.}
\label{tab:annotation-types}
\end{table}

\subsubsection{Annotators \& Procedure}
The dataset was annotated by three annotators. Two annotators are domain experts, while one annotator is a trained non-expert. 

The annotation process consisted of two stages, each followed by an annotation round.  First, the annotation scheme, theoretical background and annotation guidelines were introduced in an initial session. This was followed by the first annotation round of the complete subset. Second, a calibration session was conducted to align interpretations. This was followed by a re-annotation round of the subset of the data where labels did not agree between expert annotators. The final labels were determined via majority vote across the three annotators.

\subsubsection{Agreement \& Statistics}
The inter-annotator agreement is reported in Table \ref{tab:inter-annotator-agreement}. Results indicate substantial agreement between senior annotators and moderate agreement with junior annotator.

\begin{table}[H]
\centering
\small
\begin{tabular}{lcc}
\toprule
 & \textbf{Incivility} & \textbf{Intolerance}\\
\midrule
\textbf{Sr. 1 vs. Sr. 2} & 0.88 (0.77) & 0.90 (0.78) \\
\textbf{Sr. 1 vs. Jr. 1} & 0.71 (0.39) & 0.67 (0.37)\\
\textbf{Sr. 2 vs. Jr. 1} & 0.69 (0.38) & 0.69 (0.42)\\
\bottomrule
\end{tabular}
\caption{Agreement between senior annotators (Sr. 1 and Sr. 2) and junior annotator (Jr. 1). Table shows share of labels that agree between annotators with Cohen's $\kappa$ in brackets.}
\label{tab:inter-annotator-agreement}
\end{table}

\subsubsection{Dataset Statistics}
In total, 2{,}030 memes were annotated using the fine-grained scheme. Table~\ref{tab:marginal-distributions} reports the marginal distributions of the original coarse hatefulness label and the newly introduced incivility and intolerance labels. The statistics show that harmful content, as captured by hatefulness and intolerance, occurs at similar rates in the annotated subset.

\begin{table}[H]
\centering
\small
\begin{tabular}{lccc}
\toprule
 & \textbf{Hateful} & \textbf{Intolerant} & \textbf{Uncivil}\\
\midrule
\textbf{Original (coarse)} & 0.37 & - & -\\
\textbf{Our (fine-grained)} & 0.35 & 0.37 & 0.44\\
\bottomrule
\end{tabular}
\caption{Marginal distributions in the original dataset and our fine-grained subset.}
\label{tab:marginal-distributions}
\end{table}

Table~\ref{tab:annotation-distribution} summarizes the joint distribution of incivility and intolerance labels. The distribution highlights that incivility and intolerance frequently co-occur but also appear independently, underscoring the importance of modeling them as separate dimensions.

\begin{table}[H]
\centering
\small
\begin{tabular}{lcc}
\toprule
 & \textbf{Tolerant} & \textbf{Intolerant} \\
\midrule
\textbf{Civil} & 0.50 & 0.06 \\
\textbf{Uncivil} & 0.13 & 0.31 \\
\bottomrule
\end{tabular}
\caption{Joint distribution of incivility and intolerance labels in our fine-grained subset.}
\label{tab:annotation-distribution}
\end{table}

\section{Empirical Methodology}

\subsection{Toxicity Identification}
In hateful meme detection, harmfulness often emerges not from the text or image alone but from their interaction, requiring models to perform joint multimodal reasoning rather than isolated analysis.

To capture this interaction, we evaluate vision–language models that process visual and textual inputs in a unified framework. We test two open-source models, \textbf{LLaVA-1.6-Mistral-7B} \cite{liuVisualInstructionTuning2023} and \textbf{Qwen2.5-VL-7B} \cite{baiQwen25VLTechnicalReport2025}, both released under the Apache 2.0 license. These models represent two prominent open-source VLMs: LLaVA-1.6 as a widely adopted instruction-tuned VLM with a strong track record on vision-language benchmarks, and Qwen2.5-VL as a more recent model reflecting the current state of open-source multimodal capabilities. Both operate at the 7B parameter scale, balancing capability with the resource constraints typical of real-world moderation deployments. In addition, we evaluate \textbf{GPT-5.1}, a closed-source model accessed via the OpenAI API and governed by OpenAI’s terms of service, to approximate performance in a commercial deployment setting \cite{openai_gpt51_system_card}. It simultaneously serves as an upper bound for available commercial methods without task-specific fine-tuning.

\begin{figure}[ht]
    \centering
    \includegraphics[width=\linewidth]{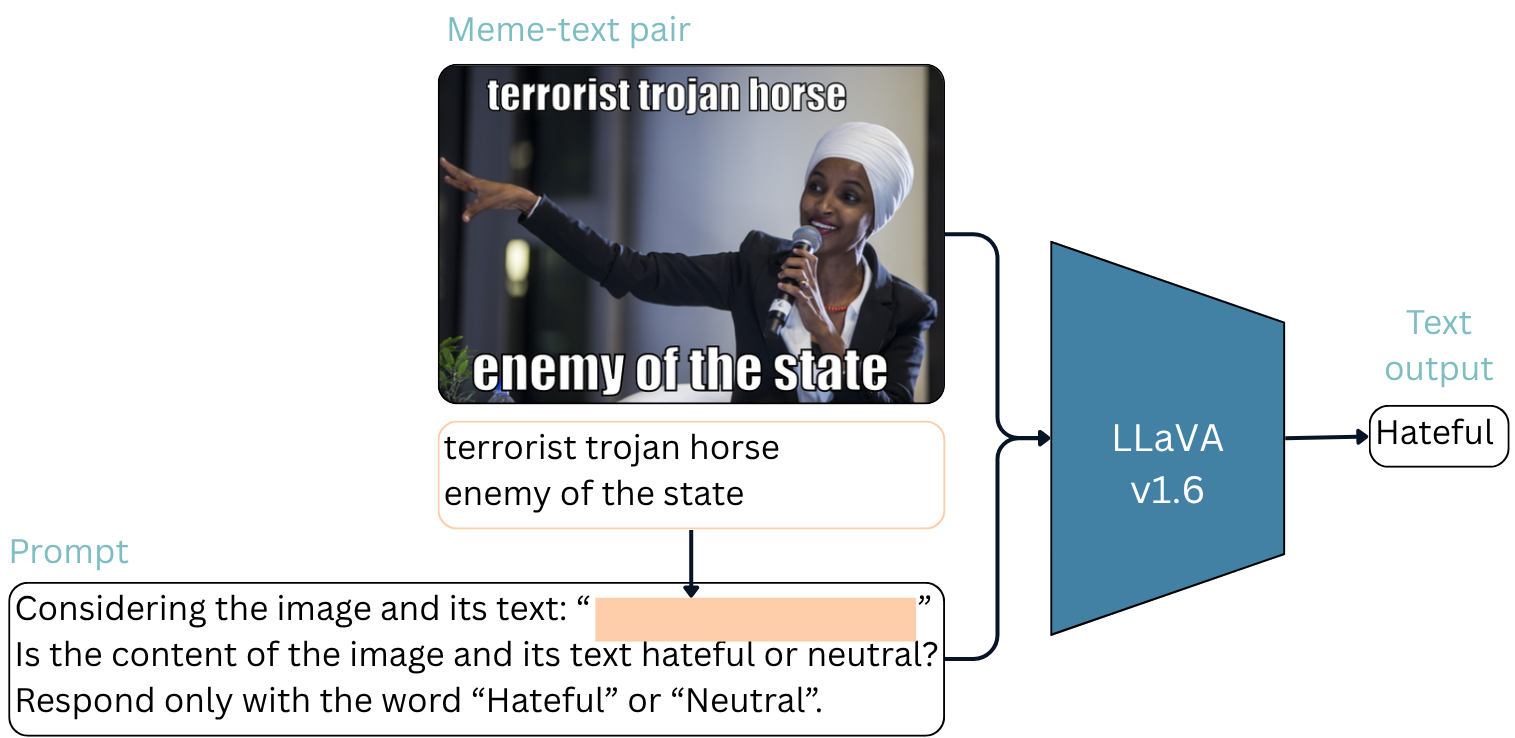}
    \caption{Evaluation pipeline. A meme–text pair is passed to the VLM (e.g. LLaVA-1.6) along with a prompt asking whether the content is ``Hateful'' or ``Neutral''. The model generates a binary classification based on both visual and textual input. Image above is a compilation of assets, including ©Getty Images.}
    \label{fig:pipeline}
\end{figure}

Vision–language models encode images into visual tokens that are combined with text tokens and jointly processed by a large language model, enabling integrated reasoning across modalities. Figure~\ref{fig:pipeline} illustrates the full experimental pipeline.

This design enables cross-modal information flow throughout the transformer layers and supports fine-grained alignment between image regions and linguistic context.

\subsection{Fine-Tuning Setups}
\label{subsec:finetuning-setups}

To evaluate both the capabilities of multimodal models and the generalisability of different annotation schemes, we consider four fine-tuning setups. These setups are designed to probe (i) off-the-shelf model competence, (ii) task-specific specialisation, (iii) cross-task generalisation, and (iv) the potential benefits of combining coarse and fine-grained supervision. Together, they allow us to assess not only model performance but also the practical implications of different labeling strategies for multimodal content moderation.

\paragraph{Overview.}
We distinguish between the following configurations:
\begin{itemize}
    \item \textbf{Zero-shot inference}, where pretrained models make 
    predictions without any task-specific fine-tuning. This setting 
    is particularly relevant for closed-source models where 
    fine-tuning is restricted, and serves as a reference point for 
    the benefit of supervised adaptation.
    \item \textbf{Baseline fine-tuning}, where models are trained and 
    evaluated using the same annotation scheme. This measures how 
    learnable each labeling strategy is given comparable supervision 
    and training budgets.
    \item \textbf{Transfer learning}, where models are trained on one 
    annotation scheme and evaluated on another, probing whether 
    representations learned under one operationalization of toxicity 
    implicitly capture information relevant to another.
    \item \textbf{Joint learning}, where models are trained 
    simultaneously on coarse and fine-grained labels, testing whether 
    the different annotation strategies provide complementary 
    supervision signals.
\end{itemize}
Each setup isolates a different aspect of model behavior and 
provides complementary evidence about how annotation granularity 
affects learning and generalisation.

\subsection{Evaluation of Fine-Tuning Setups}
We evaluate all fine-tuning setups using accuracy and weighted F1 score, standard metrics for binary classification that enable direct comparison with prior work on hate speech and toxicity detection.

\subsection{Error Biases}
Beyond overall performance, we analyze prediction errors to assess whether models exhibit systematic \textit{error biases}. In content moderation, different types of errors have different consequences: false positives correspond to over-moderation, while false negatives indicate under-detection.

We examine whether models trained under different annotation strategies tend to systematically over-moderate or under-detect harmful content. This analysis is particularly important because similar accuracy or F1 scores can mask substantially different error profiles across models and label schemes.

Our aim is to compare annotation strategies with respect to \textit{error symmetry}, that is, whether false positives and false negatives occur at comparable rates.

For each model and evaluation task, we compute the false positive rate ($FPR$) and false negative rate ($FNR$). We quantify error asymmetry as the difference between these rates, $FNR-FPR$. This measure captures whether a model is more prone to missing hateful content or to incorrectly flagging benign content. An error asymmetry of zero indicates a symmetric error profile, where false negatives and false positives occur at comparable rates. Positive values indicate under-moderation, meaning that hateful content is missed more frequently than benign content is over-flagged, whereas negative values indicate over-moderation, where benign content is more often incorrectly classified as hateful.

\subsection{Relationship Between Incivility, Intolerance, and Hatefulness}

To examine how the commonly used single-label notion of hatefulness relates to a fine-grained multi-label approach, we analyze the relationship between the original hatefulness labels in the Hateful Memes dataset and our annotations for incivility and intolerance.

The relationship between hatefulness and tone is not self-evident. While many conceptualizations treat hate as primarily content-based, benchmark definitions and annotation practice often mix semantic harm with stylistic cues such as mocking or demeaning language. Prior work has repeatedly noted ambiguity at the boundary between hate speech and merely offensive or uncivil speech. As a result, it is unclear whether the original binary hatefulness label in \emph{Hateful Memes} corresponds more closely to intolerance, incivility or their combination. We therefore evaluate four alternative operationalizations that map hatefulness to our fine-grained labels.

A meme is therefore either hateful if it is labeled as:
\begin{itemize}
    \item intolerant
    \item uncivil
    \item intolerant \textit{and} uncivil
    \item intolerant \textit{or} uncivil
\end{itemize}

For each operationalization, we compare the implied labels to the original hatefulness labels. We compute correlations to measure alignment and conduct chi-squared tests of independence to assess statistical association. This analysis clarifies whether the binary hatefulness label primarily reflects intolerance, incivility, or a combination of both.

\begin{table*}[t]
\centering
\small
\begin{tabularx}{\textwidth}{l l l *{6}{c}}
\toprule
 &  &  &
\multicolumn{2}{c}{\textbf{LLaVA-1.6-Mistral-7B}} &
\multicolumn{2}{c}{\textbf{Qwen2.5-VL-7B}} &
\multicolumn{2}{c}{\textbf{GPT-5.1}} \\
\cmidrule(lr){4-5}
\cmidrule(lr){6-7}
\cmidrule(lr){8-9}
\textbf{Setup} & \textbf{Finetune Split} & \textbf{Evaluation Split} & Acc. & F1 & Acc. & F1 & Acc. & F1 \\
\midrule

% ---------- ZERO-SHOT ----------
\multirow{3}{*}{Zero-shot}
  & - & Hateful     & 0.68 & 0.68 & 0.68 & 0.68 & \textbf{0.80} & \textbf{0.80} \\
  & - & Uncivil     & 0.61 & 0.60 & 0.56 & 0.51 & \textbf{0.74} & \textbf{0.74} \\
  & - & Intolerant  & 0.49 & 0.46 & 0.58 & 0.56 & \textbf{0.81} & \textbf{0.81} \\
\midrule

% ---------- BASELINE ----------
\multirow{3}{*}{Baseline}
  & Hateful & Hateful     & 0.73 & 0.67 & 0.77 & 0.74 & - & - \\
  & Uncivil/Intolerant & Uncivil     & 0.78 & 0.77 & 0.77 & 0.76 & - & - \\
  &                    & Intolerant  & 0.75 & 0.73 & 0.76 & 0.74 & - & - \\
\midrule

% ---------- TRANSFER LEARNING ----------
\multirow{3}{*}{Transfer Learning}
  & Uncivil/Intolerant & Hateful     & 0.73 & 0.71 & 0.75 & 0.75 & - & - \\
  & Hateful & Uncivil     & 0.74 & 0.72 & 0.70 & 0.70 & - & - \\
  &         & Intolerant  & 0.71 & 0.72 & 0.78 & 0.78 & - & - \\
\midrule

% ---------- JOINT LEARNING ----------
\multirow{3}{*}{Joint Learning}
  & \multirow{3}{*}{\makecell[l]{Uncivil/Intolerant \\ + \\ Hateful}}
    & Hateful    & \textbf{0.76} & \textbf{0.75} & \textbf{0.82} & \textbf{0.81} & - & - \\
  &              & Uncivil     & \textbf{0.79} & \textbf{0.78} & \textbf{0.82} & \textbf{0.82} & - & - \\
  &              & Intolerant  & \textbf{0.76} & \textbf{0.74} & \textbf{0.81} & \textbf{0.81} & - & - \\
\bottomrule
\end{tabularx}
\caption{Multi-model comparison for \textbf{accuracy} (Acc.) and \textbf{weighted F1 score} (F1) across all learning setups. Best results per model and evaluation split are highlighted in bold.}
\label{tab:learning-scenarios}
\end{table*}

\section{Results}
\subsection{Toxicity Identification}

The results for all models and learning setups are shown in Table~\ref{tab:learning-scenarios}. As described in the Method section, we evaluate four configurations: zero-shot inference, baseline fine-tuning, transfer learning across annotation schemes, and joint learning.

\subsubsection{Zero-shot}
In the zero-shot setting, GPT-5.1 achieves strong performance across all evaluation tasks, identifying hateful and intolerant content at similarly high rates. In contrast, open-source models show noticeably lower zero-shot performance, particularly for fine-grained labels. Both LLaVA-1.6-Mistral-7B and Qwen2.5-VL-7B achieve higher accuracy and F1 scores when predicting coarse hatefulness than when predicting intolerance or incivility, indicating that fine-grained distinctions are harder to recover without task-specific supervision.

\subsubsection{Baseline fine-tuning}
After supervised fine-tuning, both open-source models achieve broadly comparable performance across annotation schemes and evaluation metrics. Accuracy and F1 scores for coarse hatefulness and fine-grained incivility and intolerance are closely aligned, indicating that the models can successfully adapt to either labeling strategy.

One exception is observed for hatefulness prediction with LLaVA-1.6, where the F1 score is lower than in the zero-shot setting. This drop is driven by systematic under-detection of hateful content, as reflected in the high false negative rate reported in Table~\ref{tab:error_bias}. Despite this outlier, the overall results suggest that neither the coarse nor the fine-grained annotation scheme is inherently more difficult to learn, given comparable supervision and training budgets.

\subsubsection{Transfer learning}
Transfer learning results indicate substantial generalization across annotation schemes. For hatefulness and intolerance, performance differences relative to baseline fine-tuning are small and vary in direction, with minor gains in some cases and minor losses in others. These mixed outcomes do not support a clear conclusion that one annotation scheme consistently generalizes better than the other.

In contrast, incivility detection shows noticeably weaker transfer performance when models are trained solely on coarse hatefulness labels. This suggests that incivility-specific information is largely absent from single-label supervision and cannot be reliably recovered without explicit fine-grained annotation. Overall, the transfer results highlight partial semantic overlap between annotation schemes, while underscoring the limits of coarse labels for capturing stylistic dimensions such as tone.

\subsubsection{Joint learning}
Joint learning yields the strongest overall performance across tasks and models. Compared to baseline fine-tuning, jointly training on coarse hatefulness and fine-grained incivility and intolerance labels consistently improves performance for all three prediction tasks. These gains indicate that coarse and fine-grained annotation schemes are compatible and provide complementary supervision signals rather than redundant or conflicting ones. Combining both forms of annotation therefore emerges as a promising strategy for improving multimodal toxic content detection.

Under the joint learning setup, Qwen2.5-VL-7B achieves the best overall performance. It outperforms GPT-5.1 evaluated zero-shot on both hatefulness and incivility detection and matches GPT-5.1’s performance on intolerance. This result suggests that, when provided with structured and heterogeneous supervision, open-source vision–language models can reach, and even exceed the performance of strong closed-source baselines.

\subsection{Performance Biases}

When detecting harmful content (i.e. hateful or intolerant) models achieve broadly similar performance under baseline and transfer learning setups. However, comparable accuracy and F1 scores mask substantial differences in error behavior across models and annotation schemes, as shown in Table~\ref{tab:error_bias}.

GPT-5.1's error profile is not directly comparable to the fine-tuned open models, as it reflects prompt-dependent safety behavior and calibration choices inherent to commercial instruction-tuned systems. Nevertheless, the model exhibits relatively balanced false positive and false negative rates under both label schemes. The false positive rate is slightly higher than the false negative rate, indicating a mild tendency toward over-moderation that is consistent across label granularity.

In contrast, open-source models display strongly asymmetric error profiles, particularly for coarse hatefulness detection. These models are characterized by very low false positive rates but substantially higher false negative rates, reflecting systematic under-detection of harmful content. Notably, this asymmetry is reduced when models are trained and evaluated using the fine-grained intolerance labels, suggesting that finer-grained supervision can mitigate under-moderation tendencies in open-source systems.

Overall, fine-grained intolerance supervision shifts open-source models away from a strongly conservative decision rule (very low FPR but high FNR) toward a more symmetric error profile, reducing systematic under-detection of harmful content without a substantial increase in over-flagging.

\begin{table}[H]
\centering
\small
\begin{tabularx}{\columnwidth}{llccc}
\toprule
\textbf{Model} & \textbf{Task} &
\textbf{FPR} &
\textbf{FNR} &
\textbf{FNR-FPR} \\
GPT-5.1 & Hateful & 0.22 & 0.17 & \textbf{-0.05} \\
GPT-5.1 & Intolerance & 0.21 & 0.16 & \textbf{-0.05} \\
\midrule
LLaVA-1.6-Mistral-7B & Hateful & 0.01 & 0.75 & 0.74 \\
LLaVA-1.6-Mistral-7B & Intolerance & 0.09 & 0.52 & \textbf{0.42} \\
\midrule
Qwen2.5-VL-7B & Hateful & 0.04 & 0.58 & 0.54 \\
Qwen2.5-VL-7B & Intolerance & 0.09 & 0.36 & \textbf{0.28} \\
\bottomrule
\end{tabularx}
\caption{False positive rate (FPR) and false negative rate (FNR) across models and label granularity when detecting harmful content. Error asymmetry (FNR-FPR) is smaller with the fine-grained annotation scheme.}
\label{tab:error_bias}
\end{table}

\begin{table}[H]
\centering
\small
\begin{tabular}{lcc}
\toprule
\textbf{Operationalization} 
& \textbf{Match rate} 
& \textbf{Correlation} \\
\midrule
Hateful = Intolerant 
& \textbf{0.89 }
& \textbf{0.76} \\

Hateful = Intolerant AND Uncivil
& 0.88 
& 0.73 \\

Hateful = Intolerant OR Uncivil
& 0.81 
& 0.65 \\

Hateful = Uncivil 
& 0.80
& 0.59 \\

\bottomrule
\end{tabular}
\caption{Comparison of alternative operationalizations relating hatefulness to incivility and intolerance. 
Match rate indicates the proportion of cases where the label implied by a given operationalization matches the original hatefulness label. 
Correlation is the phi coefficient (equivalent to Pearson’s $r$ for binary variables). 
All associations are statistically significant according to a chi-squared test of independence at $\alpha = 0.001$.}
\label{tab:hatefulness-hypotheses}
\end{table}

\subsection{Relationship Between Incivility, Intolerance, and Hatefulness}

Table~\ref{tab:hatefulness-hypotheses} examines how the coarse hatefulness label in the Hateful Memes dataset aligns with different combinations of the fine-grained incivility and intolerance labels. For each operationalization, we compute a match rate and a correlation to assess how well the implied labels correspond to the original hatefulness annotation.

Overall, hatefulness aligns most closely with intolerance alone. The operationalization \emph{Hateful = Intolerant} yields the highest match rate (0.89) and correlation (0.76), outperforming operationalizations that additionally require incivility or rely on incivility in isolation. Requiring both intolerance and incivility (Hateful = Intolerant AND Uncivil) slightly weakens alignment, while defining hatefulness purely in terms of incivility results in substantially lower correlation.

These findings suggest that the original hatefulness label primarily captures intolerant content rather than uncivil tone. Incivility therefore represents a related but distinct dimension that is largely absent from the dataset’s operationalization of hatefulness. This supports our decision to separate tone (incivility) from content (intolerance) and shows that fine-grained annotations make explicit distinctions that remain implicit or conflated in coarse single-label schemes.

\begin{table}[h]
\centering
\small
\begin{tabular}{lccc}
\toprule
Category & n & P(hateful) & $\phi$ \\
\midrule
\multicolumn{4}{l}{\textit{Incivility}} \\
Attacks & 106 & 0.91 & 0.27 \\
Vulgar & 537 & 0.60 & 0.31 \\
Aspersions & 45 & 0.56 & 0.06 \\
Civil & 1140 & 0.10 & -0.59 \\
\midrule
\multicolumn{4}{l}{\textit{Intolerance}} \\
Threats to Rights$^\dagger$ & 6 & 1.00 & 0.07 \\
Ableism & 58 & 0.95 & 0.21 \\
Racism & 208 & 0.92 & 0.40 \\
Gender Intolerance & 153 & 0.92 & 0.34 \\
Offensive Stereotypes & 70 & 0.91 & 0.22 \\
Religious Intolerance & 186 & 0.91 & 0.37 \\
Violent Threats & 40 & 0.85 & 0.15 \\
Political Intolerance & 63 & 0.35 & -0.00 \\
Tolerant & 1278 & 0.08 & -0.76 \\
\bottomrule
\end{tabular}
\caption{Empirical hatefulness probability conditioned on fine-grained categories and $\phi$ coefficient (majority vote aggregation).}
\label{tab:granular_majority_vote}
\end{table}

To examine how different forms of incivility and intolerance relate to the original hatefulness label at a more granular level, we compute the conditional probability of hatefulness and the $\phi$ coefficient for each subtype after majority-vote aggregation (Table \ref{tab:granular_majority_vote}). Categories with fewer than 30 instances are marked with a dagger symbol and are ignored in our analysis. Full label distributions before aggregation are reported in the Appendix.

The results confirm and extend the aggregate alignment findings from Table \ref{tab:hatefulness-hypotheses}. Among intolerance subtypes, most categories exhibit hatefulness rates above 85\%, with racism ($\phi = 0.40$), religious intolerance ($\phi = 0.37$), and gender intolerance ($\phi = 0.34$) showing the strongest associations. This indicates that the coarse hatefulness label captures these forms of intolerance reliably. However, political intolerance stands out as a clear exception: despite being defined as content that delegitimizes opposing political views or calls for the elimination of political actors, it has a hatefulness rate of only 0.35 and essentially no association with the original label ($\phi = -0.00$). This suggests that the benchmark's operationalization of hate largely excludes politically-oriented intolerance, a category with particular relevance for platform governance in light of democratic discourse.

On the incivility dimension, hatefulness rates decrease along a gradient from attacks (0.91) through vulgar language (0.60) and aspersions (0.56) to civil content (0.10). However, the $\phi$ coefficients for all incivility categories remain substantially lower than those of the major intolerance subtypes, with aspersions showing near-zero association ($\phi = 0.06$). This pattern indicates that while uncivil tone frequently co-occurs with content the benchmark considers hateful, it does not independently drive the hatefulness label. The low association for aspersions is notable alongside the political intolerance finding: both suggest that the benchmark is least sensitive to forms of harmful expression that operate in the political rather than identity-based register.

%\section{Discussion}
%\lipsum[9-10]

\section{Discussion and Conclusion}
This study introduces a fine-grained annotation scheme for multimodal content moderation grounded in communication science. By explicitly distinguishing between tone (incivility) and content (intolerance), we move beyond binary notions of toxicity that dominate existing benchmarks. Our results demonstrate that fine-grained annotations complement existing coarse labels and, when used jointly, improve overall model performance. Moreover, fine-grained supervision reduces moderation-relevant error asymmetry. In our tests, open-source models shift from systematic under-detection of harmful content toward more balanced error profiles without resorting to blanket over-moderation. These gains come from improved data quality rather than model architecture, underscoring the practical value of data-centric approaches.

Our alignment analysis reveals that the original hatefulness label in the Hateful Memes dataset corresponds primarily to intolerance (match rate 0.89, $\phi = 0.76$), while incivility contributes little additional explanatory power. However, the subtype-level analysis shows that even within intolerance, the benchmark's coverage is uneven. Identity-based categories such as racism, religious intolerance and gender intolerance are captured reliably, with hatefulness rates above 90\% and strong $\phi$ coefficients. In contrast, political intolerance has a hatefulness rate of only 35\% and effectively no association with the original label ($\phi = -0.00$). Since the Hateful Memes dataset has served as the primary multimodal hate detection benchmark, these findings have implications beyond our study. Reported performance figures based on the Hateful Memes dataset should be understood as primarily reflecting identity-based intolerance detection, while politically-oriented harmful expression remains outside the benchmark's effective scope.

This selectivity is not inherently a flaw, as prioritizing identity-based intolerance may constitute a defensible moderation strategy. However, this prioritization is currently implicit, as neither the benchmark documentation nor the models trained on it signal which forms of harmful expression are included and which are excluded. Our annotation scheme makes these choices explicit by separating tone from content and preserving subtype-level distinctions, allowing system designers to see what their training data captures and what it misses. Whether to moderate political intolerance, identity-based intolerance, uncivil speech or combinations thereof is a governance decision that different platforms may reasonably answer differently and decide by conscious choice, rather than it being an artifact of benchmark construction.

\section{Limitations}
The findings of this paper should be interpreted primarily as evidence about construct operationalization and moderation-relevant error trade-offs within the Hateful Memes benchmark. Hateful Memes is deliberately curated and may not reflect the distribution, cultural context and rapidly evolving formats of memes encountered in real-world platforms. Consequently, the incidence and expression of incivility and intolerance may differ outside this setting. In addition, our fine-grained labels cover a 21\% split-stratified subset of the benchmark dataset. While learning-curve analyses suggest diminishing returns for the evaluated models at this scale, rare phenomena and less frequent subtype manifestations may be underrepresented. Finally, the annotation of incivility and intolerance remains context-sensitive despite high agreement, and our modeling focuses on binary targets, leaving subtype-level prediction to future work.

\section{Ethical Statement}
In line with the European Union's guidelines on Trustworthy AI \cite{europeancommissionEthicsGuidelinesTrustworthy2019}, this project aims to mitigate harm by detecting hateful content. We follow a data-centric approach that enhances model detection capabilities and increases transparency through the introduction of a detailed annotation codebook. This codebook provides nuanced definitions of incivility and intolerance, supporting more precise and interpretable moderation decisions.

Several ethical considerations and limitations remain. First, striking a balance between preventing harm and preserving freedom of speech is a central concern. Overly restrictive moderation risks suppressing legitimate discourse, while insufficient intervention can allow harmful content to persist. Second, assessments of content are not entirely objective; they are often shaped by the annotator’s individual perspective and sociocultural context \cite{hettiachchiHowCrowdWorker2023}, which can introduce variability and bias into annotations. Third, adaptability is a major challenge. Internet memes are fast-evolving communicative artifacts that reflect rapidly shifting cultural symbols, formats, and narratives. To remain effective, moderation systems must be resilient while also capable of adapting to emerging patterns in both form and meaning.

\section{Acknowledgements}
While drafting this paper, generative large language models were used to assist with editing, and revising portions of the manuscript. Additionally we used an AI-assisted coding tool (ClaudeCode) to support refactoring and debugging of research code. All scientific claims, experimental design decisions, data analysis, and interpretations, as well as the final code implementation, are the sole responsibility and work of the authors.

\bibliography{aaai2026}

%\clearpage
%\onecolumn
\appendix
\section{Appendix}

\section*{Appendix A: Annotation Scheme}
\label{appendix:annotation-scheme}

This appendix provides an overview of the sub-types used in our annotation scheme for incivility and intolerance. Annotators labeled each meme using the definitions below.

\subsection*{Incivility Sub-types}
To guide the annotation of incivility, we use a typology that captures different forms of uncivil expression. Each subtype is described in both text and image, since the idea of “tone” can appear differently in each. Textual incivility may involve word choice or phrasing, while visual incivility can appear through imagery or symbols. Table \ref{tab:incivility-types} lists the categories and provides a description for each modality.

\begin{table*}[t]
\centering
\small 
\begin{tabularx}{\textwidth}{l l X X}
\toprule
\textbf{Nr.} & \textbf{Form of Incivility} & \textbf{Text} & \textbf{Visual} \\
\midrule
0 & Civil & Absence of incivility. & Absence of offensive, hostile, or degrading imagery. \\
1 & Profane or vulgar & Use of explicit profanities or vulgarities, regardless of whether they are directed at a person or entity. & Images containing profanity, obscene gestures, or crude/vulgar symbolism. Sexual or violent images aimed at intimidating, shocking, or degrading public discourse. \\
2 & Attacks & Derogatory or pejorative language directed at specific individuals or groups. These can focus on personal characteristics (e.g., appearance), traits, character, or choices. & Images targeting identifiable individuals or demographic groups with pejorative, mocking, or dehumanising intent. \\
3 & Aspersions & Attacks targeting groups, organizations, or institutions aiming to undermine credibility, legitimacy, or moral standing. & Images delegitimising institutions, policies, or parties through ridicule, slurs, or hostile visual metaphors. \\
\bottomrule
\end{tabularx}
\caption{Subtypes of Incivility in Text and Visual Content. Used in the annotation scheme.}
\label{tab:incivility-types}
\end{table*}

\subsection*{Intolerance Sub-types}
To guide the annotation of intolerance, we use a typology that captures different forms of harmful content. Unlike incivility, which can manifest in different ways in text and images, the meaning of intolerant messages tends to remain consistent across modalities. Table \ref{tab:intolerance-types} presents each subtype alongside a brief description.

\begin{table*}[t]
\small 
\begin{tabularx}{\textwidth}{l l X}
\toprule
\textbf{Nr.} & \textbf{Form of Intolerance} & \textbf{Description} \\
\midrule
0 & Tolerant / Neutral & Absence of expressions of intolerance. \\
1 & Threats to Individual Rights & Claims or images where certain groups do not have equal civil, political, or human rights. \\
2 & Intolerance Toward Political Positions & Delegitimizing opposing political views or calling for elimination of political actors. \\
3 & Racism & Discriminatory, stereotypical, hateful, or prejudicial speech toward racial minorities. \\
4 & Social/Economic Intolerance & Discriminatory, stereotypical, hateful, or prejudicial speech toward others based on education level, social status, or income. \\
5 & Gender and Sexual Intolerance & Discriminatory, stereotypical, hateful, or prejudicial speech toward women and/or LGBTQ+ individuals based on gender status or sexual orientation. \\
6 & Religious Intolerance & Discriminatory, stereotypical, hateful, or prejudicial speech toward individuals or groups based on religion. \\
7 & Offensive Stereotyping & Cultural, regional, physical, or professional stereotyping with derogatory framing. \\
8 & Violent Threats & Calls to harm individuals or institutions, or support for violence. \\
9 & Ableism & Prejudice, stereotypes, mockery, exclusion, or hostility toward people with physical, intellectual, sensory, or mental disabilities. \\
\bottomrule
\end{tabularx}
\caption{Subtypes of Intolerance in Text Content. Used in the annotation scheme.}
\label{tab:intolerance-types}
\end{table*}

\section*{Appendix B: Data}
\subsection{Annotation Subset}

The original \textit{Hateful Memes} dataset contains 9{,}664 memes after removing duplicates and unavailable images. To create the subset for fine-grained annotation, we randomly sampled memes from the original dataset while stratifying by train/val/test split.

We annotated 21\% of the dataset, corresponding to 2{,}030 memes. To justify this choice, we evaluated how model performance scales with the amount of annotated data. Specifically, we fine-tuned the models for hatefulness prediction using increasing shares of the annotated subset and measured validation accuracy.

\begin{figure}[h]
    \centering
    \includegraphics[width=\linewidth]{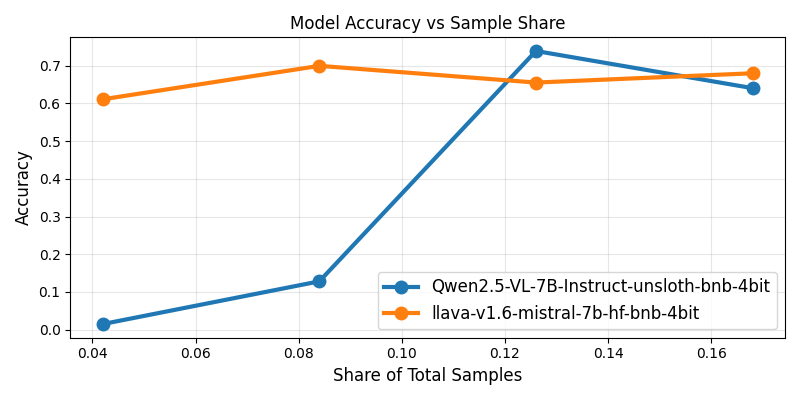}
    \caption{Validation accuracy for hatefulness prediction across different shares of annotated training data.}
    \label{fig:acc_sample_share}
\end{figure}

Figure~\ref{fig:acc_sample_share} shows accuracy as a function of the share of annotated data for both models. For LLaVA-1.6, accuracy is relatively constant across different annotation sizes. For Qwen2.5-VL-7B, accuracy increases strongly up to about 12\% of annotated data and then largely stagnates.

Overall, annotating 21\% of the dataset provides a reasonable balance between annotation effort and empirical benefit, ensuring stable training while remaining feasible for expert annotation.

\subsection{Label Frequency}

\begin{figure}[h]
    \centering
    \includegraphics[width=\linewidth]{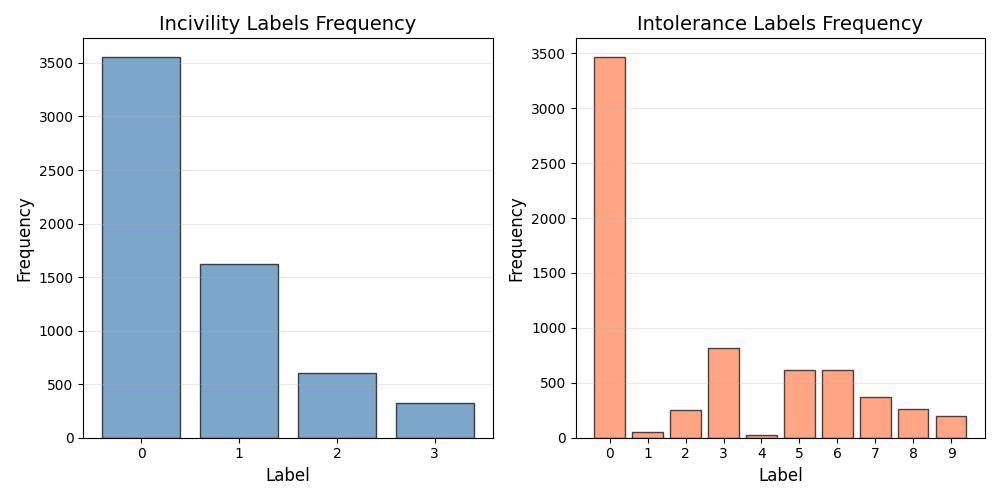}
    \caption{Frequency of fine-grained incivility (left) and intolerance (right) labels across all annotations before aggregation. A single meme may receive multiple labels.}
    \label{fig:granular_labels_frequency}
\end{figure}

Figure~\ref{fig:granular_labels_frequency} shows the frequency of each fine-grained label across all annotations before aggregation. Note that a single meme can be assigned multiple labels, as annotations capture several aspects of incivility and intolerance simultaneously.

All labels in the annotation scheme are used, but they appear with very different frequencies. For incivility, the \emph{civil} label is by far the most frequent, while more severe forms such as \emph{aspersions} occur less often. For intolerance, the \emph{tolerant} label is most common, whereas categories such as \emph{social and economic intolerance} appear comparatively rarely.

\section*{Appendix C: Training Details}
\label{appendix:training-details}

This section describes the training setup, hyperparameter tuning procedure, and computational resources used in our experiments.

\subsection*{Open-Source Models}

\paragraph{Models.}
We evaluate two open-source vision–language models: Qwen2.5-VL-7B and LLaVA-1.6-Mistral-7B. These models represent recent general-purpose VLMs with strong performance on multimodal reasoning tasks and are widely adopted in current research \cite{meiRobustAdaptationLarge2025}.

\paragraph{Model Versions.}
To reduce computational cost and improve accessibility and reproducibility, we fine-tune \textbf{quantized 4-bit} versions of both models. Quantization substantially lowers GPU memory requirements while preserving competitive downstream performance, enabling training on a single commodity GPU.

\paragraph{Fine-Tuning Strategy.}
All open-source models are fine-tuned using \textbf{Low-Rank Adaptation (LoRA)}. LoRA updates a small number of task-specific parameters while keeping the base model frozen, which significantly improves training efficiency and reduces memory consumption \cite{huLoRALowRankAdaptation2021}. This approach allows us to run multiple hyperparameter configurations within a fixed resource budget.

\subsection*{Hyperparameter Tuning}
For each task and training configuration, we perform \textbf{independent hyperparameter tuning}. All tuning runs share:
\begin{itemize}
    \item the same random-search hyperparameter space (18 different configurations),
    \item an identical tuning budget,
    \item fixed training–validation splits, and
    \item model selection based on validation \textbf{weighted F1 score}.
\end{itemize}

For reproducibility, the hyperparameter configurations corresponding to the best-performing runs are provided in Table~\ref{tab:hyperparameters}.

\begin{table*}[h]
\centering
\small
\begin{tabularx}{\textwidth}{llcccccc}
\hline
\textbf{Setup} & \textbf{Model} & \textbf{LoRA r} & \textbf{LoRA $\alpha$} & \textbf{Dropout} & \textbf{Learning Rate} & \textbf{Weight Decay} & \textbf{Epochs} \\
\hline
\multirow{2}{*}{\textbf{Coarse}} 
& Qwen2.5-VL-7B & 64 & 128 & 0.0 & 2e-3 & 0.05 & 2 \\
& LLaVA-v1.6-Mistral-7B & 64 & 128 & 0.05 & 2e-4 & 0.05 & 2 \\
\hline
\multirow{2}{*}{\textbf{Fine}} 
& Qwen2.5-VL-7B & 128 & 256 & 0.1 & 2e-3 & 0.05 & 2 \\
& LLaVA-v1.6-Mistral-7B & 64 & 128 & 0.05 & 5e-4 & 0.0 & 2 \\
\hline
\multirow{2}{*}{\textbf{Joint}} 
& Qwen2.5-VL-7B & 64 & 128 & 0.0 & 2e-3 & 0.05 & 3 \\
& LLaVA-v1.6-Mistral-7B & 128 & 256 & 0.05 & 5e-4 & 0.1 & 2 \\
\hline
\end{tabularx}
\caption{Hyperparameters for all models across training setups.}
\label{tab:hyperparameters}
\end{table*}

\subsection*{Training Results}

Table~\ref{tab:training-results} reports training results on the \textit{validation split}. For each model and learning setup, we repeat the best-performing hyperparameter configuration with five random seeds and report mean accuracy and invalid prediction rate, with the standard deviation shown in brackets.

\begin{table*}[h]
\centering
\small
\begin{tabularx}{0.9 \textwidth}{l l l cccc}
\toprule
 & & &
\multicolumn{2}{c}{\textbf{LLaVA-1.6-Mistral-7B}} &
\multicolumn{2}{c}{\textbf{Qwen2.5-VL-7B}} \\
\cmidrule(lr){4-5}
\cmidrule(lr){6-7}
\textbf{Setup} & \textbf{Finetune Split} & \textbf{Evaluation Split} & Acc. & IPR & Acc. & IPR \\
\midrule
Coarse  & Hateful & Hateful
& 0.64 ($\pm$0.05) & 0.00 ($\pm$0.00)
& 0.65 ($\pm$0.15) & 0.15 ($\pm$0.02) \\
\midrule
\multirow{2}{*}{Fine}
& \multirow{2}{*}{Uncivil/Intolerant} & Uncivil
& 0.75 ($\pm$0.02) & \multirow{2}{*}{0.00 ($\pm$0.00)}
& 0.81 ($\pm$0.03) & \multirow{2}{*}{0.05 ($\pm$0.07)} \\
&  & Intolerant
& 0.77 ($\pm$0.02) & 
& 0.76 ($\pm$0.08) &  \\
\midrule
\multirow{3}{*}{Joint}
& \multirow{3}{*}{Uncivil/Intolerant/Hateful} & Hateful
& 0.73 ($\pm$0.03) & \multirow{3}{*}{0.01 ($\pm$0.01)}
& 0.68 ($\pm$0.09) & \multirow{3}{*}{0.12 ($\pm$0.01)} \\
&  & Uncivil
& 0.75 ($\pm$0.04) &
& 0.77 ($\pm$0.02) & \\
&  & Intolerant
& 0.75 ($\pm$0.02) &
& 0.69 ($\pm$0.09) & \\
\bottomrule
\end{tabularx}
\caption{Average validation accuracy (Acc.) and invalid prediction rate (IPR) across models and training setups using 5 different seeds. Standard deviation reported in brackets.}
\label{tab:training-results}
\end{table*}

Overall, we observe relatively stable training behavior for LLaVA-1.6, as reflected by low variance across random seeds. In contrast, Qwen2.5-VL-7B exhibits higher variability across runs. This instability is primarily driven by a small number of runs producing \emph{invalid predictions}, where labels could not be reliably recovered from the generated text output. These cases introduce additional variance in the reported metrics but do not reflect systematic performance degradation of the model.

\subsection*{Computational Resources}

\paragraph{Required Resources.}
All open-source fine-tuning experiments can be executed on a single \textbf{24\,GB GPU}, made possible through 4-bit quantization and LoRA-based training.

\paragraph{Resources Used.}
Experiments were conducted on an internal compute cluster using \textbf{NVIDIA H100 94\,GB GPUs}. For the fine-grained dataset, one training epoch (1420 samples) takes approximately \textbf{17 minutes}. This setup allowed us to complete hyperparameter tuning within a practical time frame.

\subsection*{Closed-Source Model}
For comparison, we also evaluate a closed-source model using \textbf{GPT-5.1}. All inference is performed via the provider API without model fine-tuning. The total cost to evaluate the test split (407 samples) is approximately \textbf{\$ 0.30}.

\section*{Appendix D: Prompt Specifications}

All prompts follow a system–user message structure. We use three prompt variants: a coarse-grained hatefulness prompt (Listing~\ref{lst:prompt-coarse}), a fine-grained incivility and intolerance prompt (Listing~\ref{lst:prompt-fine}), and a joint prompt covering all three dimensions (Listing~\ref{lst:prompt-joint}). 

For each meme, the placeholder \texttt{\{MEME\_TEXT\}} was filled with the corresponding meme text.

\begin{listing}[h]
\caption{Coarse-grained prompt}
\label{lst:prompt-coarse}
\begin{lstlisting}
System prompt: 
You are a content moderation assistant. Aid me to label images with text as hateful or neutral. Hateful images are defined as containing a direct or indirect attack on people based on characteristics, including ethnicity, race, nationality, immigration status, religion, caste, sex, gender identity, sexual orientation, and disability or disease.
User prompt:
Considering the image and its text: {MEME_TEXT}. Is the content of the image and its text hateful or neutral?  Respond only with the word "Hateful" or "Neutral".
\end{lstlisting}
\end{listing}

\begin{listing}[h]
\caption{Fine-grained prompt}
\label{lst:prompt-fine}
\begin{lstlisting}
System prompt: 
You are a content moderation assistant. You need to classify images with text across multiple dimensions: incivility and intolerance. - Incivility: Rude, disrespectful or dismissive tone towards others as well as opinions expressed with antinormative intensity.- Intolerance: Behaviors that are threatening to democracy and pluralism - such as prejudice, segregation, hateful or violent speech, and the use of stereotyping in order to disqualify others and groups.
User prompt: 
Considering the image and its text: {MEME_TEXT}. Classify this content on two dimensions: 1. Incivility: Is this content civil or uncivil? 2. Intolerance: Is this content tolerant or intolerant? Respond in the format: "Incivility: [Civil/Uncivil], Intolerance: [Tolerant/Intolerant]"
\end{lstlisting}
\end{listing}

\begin{listing}[h]
\caption{Joint prompt}
\label{lst:prompt-joint}
\begin{lstlisting}
System prompt: 
You are a content moderation assistant. You need to classify images with text across multiple dimensions: incivility, intolerance, and hatefulness. - Incivility: Rude, disrespectful or dismissive tone towards others as well as opinions expressed with antinormative intensity.- Intolerance: Behaviors that are threatening to democracy and pluralism - such as prejudice, segregation, hateful or violent speech, and the use of stereotyping in order to disqualify others and groups.- Hatefulness: Hateful content is defined as containing a direct or indirect attack on people based on characteristics, including ethnicity, race, nationality, immigration status, religion, caste, sex, gender identity, sexual orientation, and disability or disease.
User prompt: 
Considering the image and its text: {MEME_TEXT}. Classify this content on three dimensions: 1. Incivility: Is this content civil or uncivil? 2. Intolerance: Is this content tolerant or intolerant? 3. Hatefulness: Is this content hateful or neutral? Respond in the format: "Incivility: [Civil/Uncivil], Intolerance: [Tolerant/Intolerant], Hatefulness: [Hateful/Neutral]"
\end{lstlisting}
\end{listing}

\section* {Appendix E: Additional Annotation Analyses}

\subsection{Raw fine-grained frequencies}
This section contains data on raw fine-grained frequencies. Table \ref{tab:incivility_finegrained_distribution} contains the distribution for incivility. Table \ref{tab:intolerance_finegrained_distribution} contains the distribution for intolerance. Note that these represent all labels. A large subset of data points received multiple valid labels in these categories.

\begin{table}[H]
\centering
\small
\begin{tabular}{lcc}
\toprule
\textbf{Category} & \textbf{Count} & \textbf{Share}\\
\midrule
\textbf{Civil} & 3552 & 0.58\\
\textbf{Vulgar} & 1623 & 0.27\\
\textbf{Attacks} & 603 & 0.10\\
\textbf{Aspersions} & 323 & 0.05\\
\midrule
\textbf{Total} & \textbf{6101} & \textbf{1.00}\\
\bottomrule
\end{tabular}
\caption{Distribution of incivility categories in the dataset.}
\label{tab:incivility_finegrained_distribution}
\end{table}

\begin{table}[H]
\centering
\small
\begin{tabular}{lcc}
\toprule
\textbf{Category} & \textbf{Count} & \textbf{Share}\\
\midrule
\textbf{Tolerant} & 3463 & 0.52\\
\textbf{Threats to Rights} & 56 & 0.01\\
\textbf{Political Intolerance} & 255 & 0.04\\
\textbf{Racism} & 815 & 0.12\\
\textbf{Social Intolerance} & 25 & 0.00\\
\textbf{Gender Intolerance} & 615 & 0.09\\
\textbf{Religious Intolerance} & 615 & 0.09\\
\textbf{Offensive Stereotypes} & 372 & 0.06\\
\textbf{Violent Threats} & 259 & 0.04\\
\textbf{Ableism} & 195 & 0.03\\
\midrule
\textbf{Total} & \textbf{6670} & \textbf{1.00}\\
\bottomrule
\end{tabular}
\caption{Distribution of intolerance categories in the dataset.}
\label{tab:intolerance_finegrained_distribution}
\end{table}

\subsection{Fine-grained labels majority vote}

\begin{figure}[H]
    \centering
    \includegraphics[width=\linewidth]{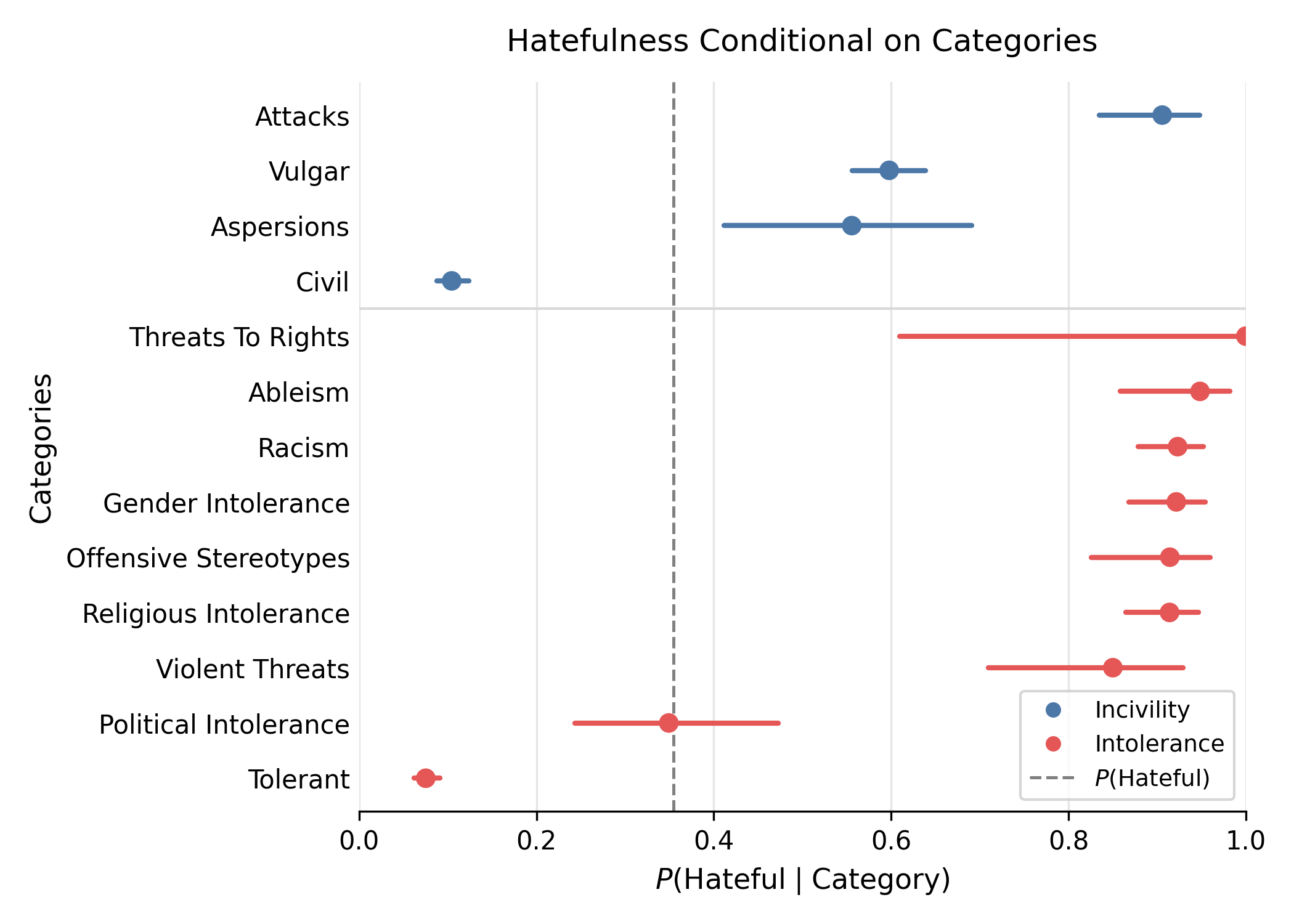}
    \caption{Hatefulness empirical probability conditioned on each category. Error bars represent the Wilson Score Interval at 0.05 significance level.}
    \label{fig:hateful_cond_granular_cat}
\end{figure}

In addition to the probabilities shown in the main part of the paper, we present the conditional probabilities for the majority voting on different categories in Figure \ref{fig:hateful_cond_granular_cat}. This includes the Wilson Score Interval in the form of error bars, beyond the numbers represented in the Table from the main paper. Note that Threats to Rights is again a special case, due to the low number of memes that received this annotation.

\subsection*{Logistic Regression on Fine-Grained Labels}

Performing a naive majority-vote aggregation of the fine-grained incivility and intolerance labels discards a large part of the annotations where there is no clear majority category across annotators.
To prevent this information loss, we conducted additional experiments, where we represent each annotator's response as a
probability distribution over the label categories, where the weight of $1$ is split
equally across all categories selected by that annotator.

To assess how these granular labels relate to hatefulness, we fit a logistic regression
with the aggregated category probabilities as predictors and the majority-voted binary
hateful label as the outcome.
The model is stable and robust: it achieves a high $R^2$ ($0.60$), a significant
log-likelihood ratio $p$-value ($< 0.000$), and converges after $7$ iterations.
The estimated intercept implies that only $\exp(-4.2) \approx 1.4\%$ of memes coded as
both civil and tolerant are labeled hateful, confirming that the baseline rate for
non-offensive content is near zero.

\begin{figure}[H]
    \centering
    \includegraphics[width=\linewidth]{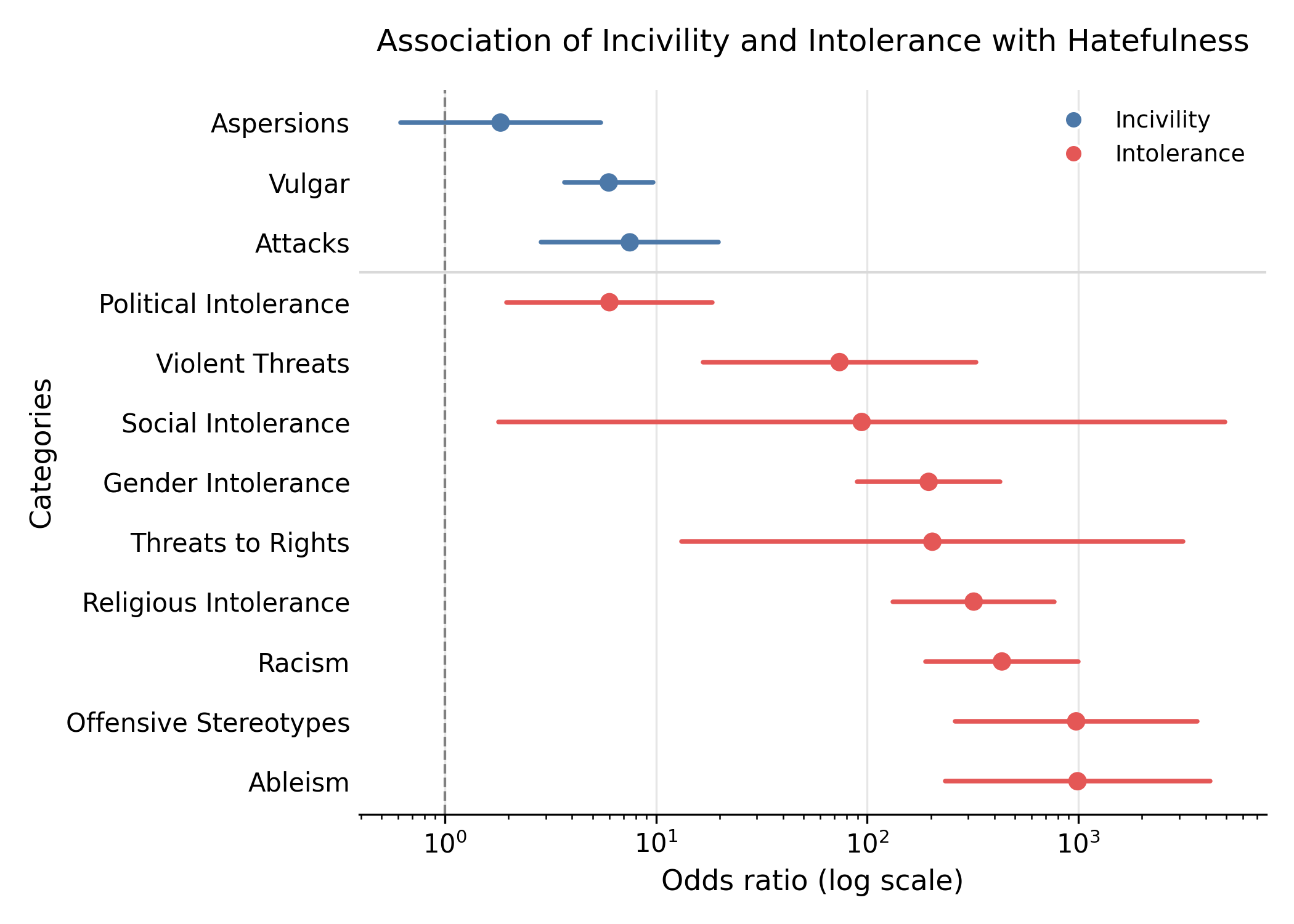}
    \caption{Odds ratios from a logistic regression of the hateful label on fine-grained incivility and intolerance category probabilities. Points to the right of the dashed line ($\mathrm{OR} > 1$) indicate a positive association with hatefulness. Error bars denote 95\% confidence intervals.}
    \label{fig:regression_odds_ratios}
\end{figure}

The results from this regression again reveal a structural difference between incivility and intolerance.
Incivility categories (with the exception of aspersions) significantly increase the
odds of hatefulness, but the effect sizes are moderate.
Intolerance categories, by contrast, are associated with substantially higher odds
ratios, indicating that explicit ideological targeting is a much stronger signal of
hatefulness than linguistic incivility alone.
Among individual categories, \emph{Racism}, \emph{Offensive Stereotypes}, and
\emph{Ableism} emerge as the clearest predictors of hateful content.

\end{document}